\documentclass{article}

\PassOptionsToPackage{numbers, compress}{natbib}
\usepackage[sglblindworkshop, final]{neurips_2026}
\workshoptitle{Towards Test-Time Continual Learning Agents}

\workshoptitle{Towards Test-Time Continual Learning Agents}

\usepackage[utf8]{inputenc} 
\usepackage[T1]{fontenc}    
\usepackage{hyperref}       
\usepackage{url}            
\usepackage{booktabs}       
\usepackage{amsfonts}       
\usepackage{nicefrac}       
\usepackage{microtype}      
\usepackage{xcolor}         
\usepackage{wrapfig}        
\usepackage{enumitem}
\usepackage{amsthm}
\newtheorem{proposition}{Proposition}
\usepackage{booktabs}
\usepackage{multirow}
\usepackage{bm}
\usepackage{amsthm}
\usepackage{multirow}
\usepackage{algorithm}
\usepackage{algorithmic}
\usepackage{mathtools}
\usepackage{amssymb}
\DeclarePairedDelimiterX{\norm}[1]{\lVert}{\rVert}{\mathopen{}\mathclose\bgroup#1\egroup}

\usepackage{caption}
\usepackage[T1]{fontenc}
\usepackage[scaled=0.92]{inconsolata}
\usepackage{xcolor}
\usepackage{listings}
\usepackage{svg}
\usepackage{cleveref} 
\definecolor{plasmaA}{HTML}{6A00A8}
\definecolor{plasmaB}{HTML}{B12A90}
\definecolor{plasmaC}{HTML}{E16462}
\definecolor{plasmaBG}{rgb}{0.98,0.98,0.98} 

\newcommand{\highlight}[1]{\textbf{\textcolor{plasmaB}{#1}}}
\newcommand{\highlightA}[1]{\textbf{\textcolor{plasmaA}{#1}}}
\newcommand{\highlightB}[1]{\textbf{\textcolor{plasmaC}{#1}}}

\lstdefinestyle{pyplasma}{
  language=Python,
  basicstyle=\ttfamily\small,
  numbers=left,
  numberstyle=\color{black}\scriptsize, 
  numbersep=7pt,                        
  xleftmargin=1.5em,
  showstringspaces=false, upquote=true,
  columns=fullflexible, keepspaces=true,
  breaklines=true, breakatwhitespace=false, breakindent=1.5em,
  postbreak=\mbox{\textcolor{gray}{$\hookrightarrow$}\space},
  literate=* {.}{{.}}1 {(}{{(}}1 {)}{{)}}1 {[}{{[}}1 {]}{{]}}1
            {/}{{/}}1 {=}{{=}}1 {,}{{,}}1,
  alsoletter={_},
  moredelim=**[is][\bfseries\color{plasmaB}]{\\highlight\{}{\}},
  moredelim=**[is][\bfseries\color{plasmaA}]{\\highlightA\{}{\}},
  moredelim=**[is][\bfseries\color{plasmaC}]{\\highlightB\{}{\}},
}

\title{Decoupling Memory from Context: Structured Memory for Token-Efficient Test-Time Continual Learning}

\author{%
  Yehya Farhat \\ 
   Rice University,\ Department of Computer Science\\
   Houston, TX, USA \\
  \texttt{Yehya.Farhat@rice.edu} \\
   \And
   Michael Desmond \\
  IBM Research \\
  Yorktown Heights, NY, USA \\
   mdesmond@us.ibm.com \\
   \AND
  Anastasios Kyrillidis \\
  Rice University, Department of Computer Science \\
   Houston, TX, USA \\
  anastasios@rice.edu\\
}

\begin{document}

\maketitle

\begin{abstract}
Large language models (LLMs) are increasingly deployed in enterprise, scientific, and medical applications, where agents must incorporate domain-specific knowledge and adapt from experience. Context engineering offers a practical alternative to weight updates by improving model behavior through instructions, strategies, and evidence supplied at inference time. However, adapting context online typically requires a costly trial-and-error process, while queries are often processed independently, preventing useful experience from carrying forward. Memory systems address this limitation by retaining information across interactions, but approaches that continually append information to a shared context face increasing token costs, context-window limits, and performance degradation as the context expands. We introduce a unified formulation of context optimization and show that an agent memory system update can be interpreted as optimization update procedure over the model’s context. This perspective attempts to provide a principled framework for studying memory design and its efficiency. We then propose GraphMemory, a lightweight graph-based memory that accumulates, refines, organizes, and connects reusable strategies. For each query, GraphMemory retrieves only the relevant subgraph, enabling online context adaptation without exposing the model to the entire memory. Under bounded retrieval, the amount of retrieved memory remains constant as the number of processed examples grows. Experiments show that GraphMemory achieves competitive downstream performance while using approximately $81$--$85\%$ fewer memory-construction tokens than our baselines.

\end{abstract}

\section{Introduction}
\vspace{-0.9em}

Modern large language models (LLMs) achieve strong performance across reasoning, question answering, and open-ended generation tasks \cite{xi2025rise}. Yet deployed agents largely remain static: with fixed parameters, information acquired from one interaction does not naturally improve later behavior. Parameter-efficient fine-tuning (PEFT) provides one mechanism for adaptation \cite{hu2021lora, dun2025sweepingheterogeneitysmartmops, li2021prefix}, while more recent methods generate context-specific adapters on the fly in a single forward pass \cite{charakorn2026doctoloralearninginstantlyinternalize}. These approaches still require access to model weights and training infrastructure, and encode their learned state in model-specific parameters. This limits continual adaptation during deployment and makes the acquired knowledge difficult to transfer across agents.

Context engineering offers a weight-free alternative by adapting a frozen model through instructions, demonstrations, strategies, or evidence supplied alongside each query to leverage its in-context learning abilities \cite{brown2020language}. Prompt-optimization methods treat this context as the optimization variable and search for task-level prompts that improve performance \cite{yang2024largelanguagemodelsoptimizers, khattab2023dspycompilingdeclarativelanguage, opsahlong2024optimizinginstructionsdemonstrationsmultistage, agrawal2026gepareflectivepromptevolution}. However, they generally optimize a single prompt against an offline dataset or user-defined evaluator rather than accumulating experience during deployment. Other methods use a hybrid setup that derives training signals from model behavior itself \cite{ramji2026latent}. 
Persistent memory systems instead update an external natural-language memory as new problems are solved. Dynamic Cheatsheet stores reusable strategies in a persistent file \cite{suzgun2025dynamiccheatsheettesttimelearning}, while Agentic Context Engineering (ACE) incrementally evolves a structured playbook through a Generator--Reflector--Curator architecture \cite{zhang2026agenticcontextengineeringevolving}. Because this memory remains external to the model, it can be inspected, edited, and transferred across agents and has shown success in large-scale enterprise applications \cite{tang2026wikiskill, ni2026trace2skill}. 

Nevertheless, scaling such memory introduces a fundamental problem. Because existing systems largely maintain a flat collection or a single evolving file, the model must process an increasingly large and heterogeneous memory context for every query. This increases token usage, pushes against context-window limits, and can lead to context degradation or collapse. To address these limitations, we introduce \textsc{GraphMemory}, a structured external memory for scalable test-time learning. \textsc{GraphMemory} represents reusable ACE-style knowledge points as nodes and captures useful relationships between them as weighted edges. For each query, the system identifies a small set of relevant entry nodes and traverses their local neighborhoods, exposing only the resulting subgraph to the language model. This representation enables selective retrieval without requiring all experience to be compressed into one global playbook or assigned exclusively to an isolated section. It also makes relationships among individual lessons explicit, allowing knowledge acquired from different experiences to be retrieved together when their connection is useful to the current problem. Our contributions are listed as follows:

\begin{itemize}
\item \textbf{A unified context-optimization framework.}
We formulate adaptation of a frozen language model as optimization over its textual context and show that persistent memory systems can be interpreted as optimization procedures that update this context across examples. Under bounded update sizes, we show that the context used by append-only memory grows as $O(N)$, whereas bounded retrieval maintains an $O(1)$ memory-context length per query.

\item \textbf{GraphMemory.}
We introduce \textsc{GraphMemory}, a lightweight external memory that accumulates, refines, organizes, and connects reusable strategies in an evolving graph. Nodes represent ACE-style knowledge points, while weighted edges capture useful relationships among them. For each query, GraphMemory identifies relevant entry nodes, traverses their local neighborhoods, and supplies only the resulting subgraph to the language model. This replaces full-memory prompting with selective relational retrieval while preserving the interpretability, portability, and weight-free adaptation of natural-language memory.

\item \textbf{Competitive performance at substantially lower cost.}
Across three dataset--model settings on two financial reasoning tasks, \textsc{Formula} and \textsc{FiNER}, GraphMemory remains within $4.0$ percentage points of ACE with Claude Haiku 4.5 and outperforms ACE by $1.5$ points on \textsc{Formula} with Qwen3.8-27B. GraphMemory reduces training-token usage by $81$--$85\%$, test-token usage by $79$--$92\%$, and active training time by $21$--$34\%$. 

\end{itemize}

\section{Context and Memory Formalization}
\vspace{-0.9em}

\subsection{Offline Context Optimization}

Let $\mathcal{D}=\{(x_i,y_i)\}_{i=1}^{n}$ be an offline training set, let
$f$ be a frozen language model, and let $c\in\mathcal{C}$ denote textual
context supplied alongside an input. We define the empirical context risk as
\begin{equation}
    L_{\mathcal D}(c)
    = \frac{1}{n}\sum_{i=1}^{n}
    \mathbb{E}\left[
        \ell\bigl(f(c,x_i),y_i\bigr)
    \right],
    \label{eq:context-risk}
\end{equation}
where the expectation is over any randomness in the model or its execution
environment. Context optimization seeks
$c^*\in\arg\min_{c\in\mathcal C} L_{\mathcal D}(c)$. The context may
contain instructions, demonstrations, knowledge, strategies, or any
combination thereof. Because $c$ is a discrete sequence of tokens and the
model may be accessible only through a black-box interface, ordinary
first-order optimization with respect to $c$ is unavailable.

Prompt-optimization methods such as MIPROv2 and GEPA
\cite{opsahlong2024optimizinginstructionsdemonstrationsmultistage,
agrawal2026gepareflectivepromptevolution} approximate this problem by
restricting the search to proposed candidates. Let $\mathcal P_T\subset\mathcal C$
denote the candidate pool considered within budget $T$. Their shared abstraction is
\begin{equation}
    \widehat c_T
    \in \arg\min_{c\in\mathcal P_T}
    L_{\mathcal D}(c).
    \label{eq:finite-context-search}
\end{equation}
The methods differ in how $\mathcal P_T$ is constructed and searched. When
abstracted to a single context variable, Equation~\eqref{eq:finite-context-search}
produces a task-level context that is held fixed on the test set. We give the
more specific formulations in Appendix~\ref{app:optimization-oracles}.

A complementary family of methods uses model feedback to iteratively propose
textual changes to the current context. Let $\widehat y_t=f(c_t,x_t)$, and let
$\tau_t$ contain the available feedback and execution trace. We abstract such
an update as
\begin{equation}
    \Delta c_t
    = \mathcal O(c_t,x_t,\widehat y_t,\tau_t),
    \qquad
    c_{t+1}=c_t\odot\Delta c_t,
    \label{eq:textual-context-update}
\end{equation}
where $\mathcal O$ is a black-box optimization oracle and $\odot$ applies the
proposed addition, deletion, replacement, or edit.
Equation~\eqref{eq:textual-context-update}
is structurally analogous to a gradient-descent step over the context:
$\Delta c_t$ plays the role of an update direction. It is not a literal
gradient, however, and it need not decrease Equation~\eqref{eq:context-risk}.
Methods such as OPRO, TextGrad, and TRACE \cite{cheng2024trace, yuksekgonul2024textgrad, yang2024large} differ primarily in how they
construct the oracle $\mathcal{O}$ and the update. We discuss these realizations in
Appendix~\ref{app:optimization-oracles}.

\subsection{From Local Context Updates to Memory}

Directly asking an imperfect oracle to improve one global context over the
entire dataset is difficult. Memory systems instead extract a short context
block $\Delta_i$ from an individual experience $(x_i,y_i)$. Writing the collection of context blocks as
$\delta=(\Delta_1,\ldots,\Delta_n)$, we define the context-block risk
as the local surrogate
\begin{equation}
    L^{\mathrm{block}}_{\mathcal D}(\delta)
    = \frac{1}{n}\sum_{i=1}^{n}
    \mathbb{E}\left[
        \ell\bigl(f(\Delta_i,x_i),y_i\bigr)
    \right].
    \label{eq:context-block-risk}
\end{equation}
This surrogate decomposes update generation into smaller problems: the oracle
need only distill a focused lesson, for example, a strategy, correction, or
failure mode, from one experience at a time. The blocks collectively form an
external memory bank $\delta$, which serves as the evolving optimization state
of the memory system.

Optimizing Equation~\eqref{eq:context-block-risk} is not claimed to be equivalent to optimizing the
global objective in Equation~\eqref{eq:context-risk}. A block may overfit its
source example and need not generalize in isolation. The motivation for this surrogate is primarily practical. First, it models the update procedure of a memory system. Second, it makes the oracle's task more local: proposing a focused update $\Delta_i$ is a simpler generation task than proposing changes to improve performance over the entire dataset. A separate composition or retrieval mechanism
must determine which blocks should form the context for a new input; the local surrogate alone does not ensure generalization. We expand
on this distinction in Appendix~\ref{app:local-context-objective}. Dynamic Cheatsheet and ACE instantiate this view by integrating locally
extracted lessons into a shared natural-language memory
\cite{suzgun2025dynamiccheatsheettesttimelearning,
zhang2026agenticcontextengineeringevolving}. The blocks are integrated into one shared context, making accumulated lessons
available across examples and producing a single task-level memory for deployment. The resulting memory must therefore serve new inputs rather than supply example-indexed solutions.

Under an append-only abstraction, this context
evolves as
\begin{equation}
    c_{\delta}^{(t+1)}
    = c_{\delta}^{(t)}\oplus\Delta^{(t)},
    \label{eq:append-only-memory}
\end{equation}

where $\oplus$ appends the new block to the shared context $c_{\delta}$. This is an idealized regime: Dynamic
Cheatsheet and ACE may edit, merge, or delete existing content rather than
literally appending every update. Nevertheless, these operations do not
preclude systematic context growth. In
Figure~\ref{fig:theory_vs_measured}, we show that the measured growth of ACE's
prompt-token usage is consistent with the growth trend predicted by the update regime in Equation~\eqref{eq:append-only-memory} over the evaluated training horizon.

\begin{proposition}[Append-only memory-context growth]
\label{prop:append-only-growth}
Suppose that $N$ examples produce at most one block each and that
$\operatorname{len}(\Delta^{(t)})\leq B$, where $B$ is independent of $N$.
Under Equation~\eqref{eq:append-only-memory}, with a fixed initial context, the final memory-context length
satisfies
\begin{equation}
    \operatorname{len}\left(c_{\delta}^{(N)}\right)
    \leq \operatorname{len}\left(c_{\delta}^{(0)}\right)+NB
    = O(N).
    \label{eq:append-only-growth-bound}
\end{equation}
\end{proposition}
The proof follows by unrolling the induced length recurrence and is given in
Appendix~\ref{app:context-scaling-proofs}.

This growth matters because the shared memory is not stored and read only
once: it is repeatedly supplied to the model on subsequent examples. The
amount of memory processed per query therefore increases with the number of
accumulated time-steps (lessons). Across $N$ sequential updates, this can produce
quadratic cumulative memory-context exposure, even though the final context
itself grows only linearly. In practice, the expanding context also moves the
system toward the model's context-window limit and mixes an increasingly
heterogeneous collection of lessons, only a fraction of which may be relevant to the current input. The corresponding cumulative bound is derived in  Appendix~\ref{app:context-scaling-proofs}.
The resulting efficiency problem is not that the stored memory grows; a
continual learner should retain experience. The problem is that full-memory
systems couple stored-memory size to the amount of memory processed on every
query. This motivates a memory interface that allows the stored knowledge to
continue growing while keeping the active context small.

\subsection{Memory with Bounded Retrieval}

A more token-efficient procedure introduces an interface that separates the accumulated memory $\delta$ from the context $c$ exposed to the model:
\begin{equation}
    \delta^{(t+1)}
    = \delta^{(t)}\cup\{\Delta^{(t)}\},
    \qquad
    c_{\phi}^{(t+1)}(x)
    = \phi\bigl(\delta^{(t+1)},x\bigr),
    \label{eq:retrieval-memory-update}
\end{equation}
where $\phi$ retrieves a query-dependent subset of the accumulated blocks.
The oracle $\mathcal O$ determines what is learned from each experience, while
$\phi$ determines which learned information is exposed for a particular
input.

To this end, we introduce GraphMemory and model $\delta$ as a graph $G^{(t)}=(V^{(t)},E^{(t)})$. Each
node stores a reusable context block $\Delta$, and weighted edges identify blocks that
may be useful together. The graph is a modeling choice for instantiating $\phi$ that supports selective retrieval and composition of related blocks. Our framework also admits other representations: for example, each block in $\delta$ could be embedded and $\phi$ implemented as a similarity-based retriever. In GraphMemory, for each query, $\phi$ selects entry nodes from a
condensed index and expands through their neighborhoods under a fixed budget; only the retrieved subgraph is supplied to the model $f$. The complete graph
update and its relation to full-memory exposure are given in Appendix~\ref{app:graph-formalization}; the concrete retrieval and update
procedures are described in the following section.

\begin{proposition}[Bounded context under retrieval]
\label{prop:bounded-retrieval}
Suppose each stored block contains at most $B$ tokens and $\phi$ retrieves at
most $K$ blocks, where $B$ and $K$ are independent of $N$. Then, for every
query $x$, the retrieved memory context satisfies
\begin{equation}
    \operatorname{len}\left(c_{\phi}^{(N)}(x)\right)
    \leq KB
    = O(1) 
    \label{eq:bounded-retrieval-context}
\end{equation}
\end{proposition}

\begin{figure}[t]
  \centering
  \begin{minipage}[t]{0.7\linewidth}
    \centering
    \includegraphics[width=\linewidth]{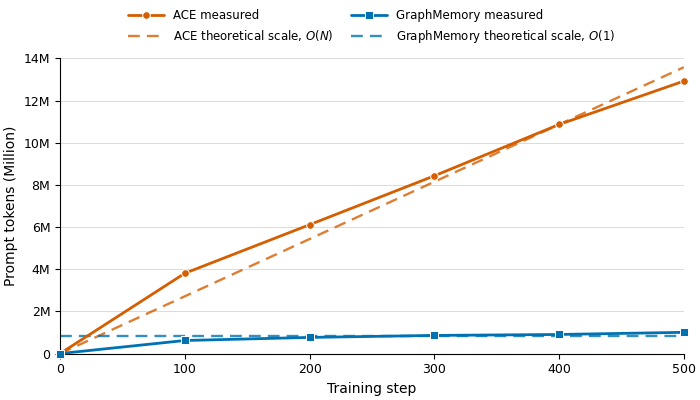}
    \caption{Prompt-token usage at successive training checkpoints for GraphMemory and ACE on \textsc{Formula} with Qwen3.8-27B.}
    \label{fig:theory_vs_measured}
  \end{minipage}
\end{figure}

The proof and cumulative memory-context bound are given in
Appendix~\ref{app:context-scaling-proofs}. Proposition~\ref{prop:bounded-retrieval}
concerns only the memory supplied to the Generator; it does not bound storage or the internal cost of retrieval. Figure~\ref{fig:theory_vs_measured} illustrates the different prompt-token growth trends of ACE and GraphMemory over the evaluated training horizon.

\section{GraphMemory}
\vspace{-0.9em}
\label{sec:GraphMemory}

\begin{figure}[t]
\centering
\includegraphics[width=0.9\linewidth]{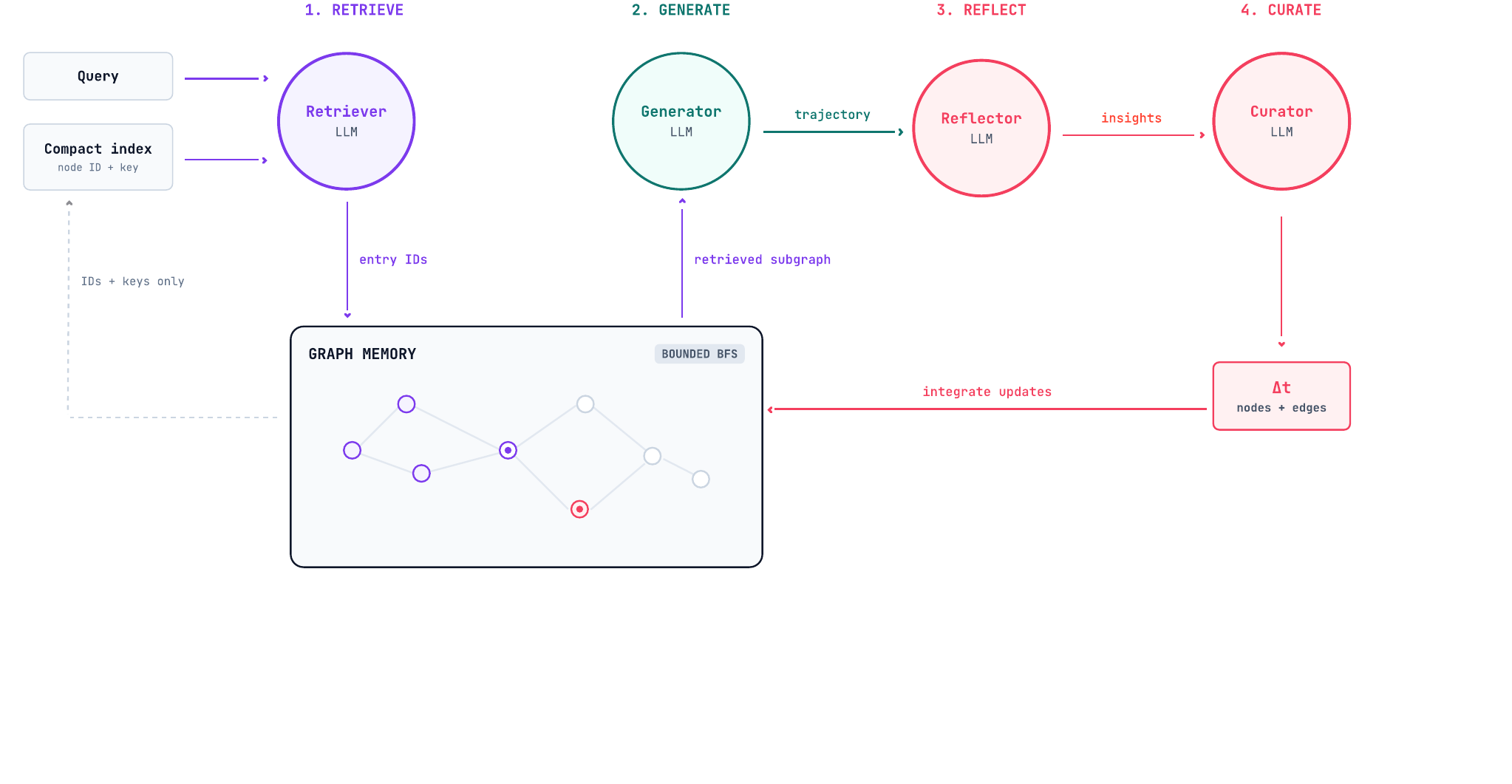}
\vspace{-2em}
\caption{
Overview of GraphMemory. An LLM retriever selects entry nodes from a
compact index, after which a bounded breadth-first traversal loads the
relevant subgraph. The ACE Generator and Reflector operate on this
retrieved context, while the Curator writes new nodes and edges back
to the graph.
}
\label{fig:GraphMemory-pipeline}
\end{figure}

We use ACE as our workhorse for realizing GraphMemory. GraphMemory changes ACE at precisely two interfaces: what the Generator ($f$) reads and what the Curator ($\mathcal{O}$) writes. Instead of exposing the complete playbook to every query, we prepend a retrieval stage; instead of writing only bullet-point updates, the Curator can also write relations between them. The resulting adaptation loop is \textsc{Retrieve--Generate--Reflect--Curate}, as illustrated in Figure~\ref{fig:GraphMemory-pipeline}. For retrieval, we maintain a compact index containing each node's identifier and a short key, but not its full bullet-point content. Given a query, the Retriever reads this index and selects a small set of entry nodes. We then perform a deterministic breadth-first traversal from those entries, expanding neighbors in order of learned edge utility, under fixed hop, node, and token budgets. Only the full contents of the nodes reached during this traversal are loaded and passed to the Generator. The LLM therefore makes the initial semantic choice, while the graph provides a cheap and deterministic mechanism for bringing in connected knowledge. 

Generation follows ACE. Reflection is extended in one respect: beyond ACE-style critique, the Reflector labels the Generator's reported memory transitions as helpful, harmful, or neutral. Helpful and harmful labels contribute confidence-weighted evidence to the corresponding edges, while neutral labels leave the counters unchanged. The traversal utility is the mean of the resulting Beta posterior, so retrieval preferentially follows transitions with stronger helpful evidence. The difference appears during curation: the Curator produces a graph update that may add or revise bullet-point nodes, add or revise edges, or do both. The update is applied locally without rewriting the rest of the graph. New nodes are added to the compact index, while new edges determine which lessons can be reached together during later retrieval. We provide the detailed construction of GraphMemory in Appendix~\ref{app:implementation-details}. It is important to note that this realization is not strictly $O(1)$ end to end because the Retriever still reads an index that grows with the number of nodes. The bound in Proposition~\ref{prop:bounded-retrieval} applies to the full memory content supplied to the Generator. Since each index entry contains only an identifier and a short key, however, this residual growth is far smaller than repeatedly supplying every bullet in full. Over the training horizon studied here, the resulting context grows slowly enough to exhibit the near-constant behavior shown in Figure~\ref{fig:theory_vs_measured}.

\section{Experiments}
\vspace{-0.9em}

\begin{table*}[t]
    \centering
\caption{
    Offline adaptation performance, token usage, and estimated training
    cost on \textsc{Formula} and \textsc{FiNER}. Training-token counts and costs are approximate.
    Costs assume a $95\%-5\%$ input--output token split and standard
    real-time API prices. Initial accuracy denotes cold-start test accuracy.
\textsuperscript{$\dagger$} ACE/Qwen on \textsc{FiNER} is not reported; see note below.
    }
    \label{tab:combined-results}
    \setlength{\tabcolsep}{4pt}

    \resizebox{\textwidth}{!}{%
    \begin{tabular}{@{}lllcccrrr@{}}
        \toprule
        \textbf{Dataset}
        & \textbf{Backbone}
        & \textbf{System}
        & \multicolumn{3}{c}{\textbf{Accuracy}}
        & \multicolumn{2}{c}{\textbf{Token Usage}}
       & \multicolumn{1}{c}{\textbf{Estimated Cost}\textsuperscript{*}} \\
        \cmidrule(lr){4-6}
        \cmidrule(lr){7-8}
        \cmidrule(l){9-9}
        &
        &
        & \textbf{Initial}
        & \textbf{Best val.\ (step)}
        & \textbf{Final test}
        & \textbf{Training}
        & \textbf{Test}
        & \textbf{Training (USD)} \\
        \midrule

        \multirow{4}{*}{\textsc{Formula}}
        & \multirow{2}{*}{Claude Haiku 4.5}
        & GraphMemory
        & 0.70
        & 0.87 (500)
        & 0.81
        & \textbf{$\sim$20M}
        & \textbf{$\sim$1.4M}
        & \textbf{\$24.00} \\

        &
        & ACE (80k cap)
        & 0.70
        & \textbf{0.88 (300)}
        & \textbf{0.85}
        & $\sim$131M
        & $\sim$18M
        & \$157.20 \\

        \cmidrule(l){2-9}

        & \multirow{2}{*}{Qwen3.8-27B}
        & GraphMemory
        & 0.69
        & \textbf{0.88 (200)}
        & \textbf{0.84}
        & \textbf{$\sim$8M}
        & \textbf{$\sim$0.6M}
        & \textbf{\$5.00} \\

        &
        & ACE
        & 0.67
        & 0.88 (200)
        & 0.83
        & $\sim$42M
        & $\sim$4M
        & \$26.25 \\

        \midrule

        \multirow{2}{*}{\textsc{FiNER}}
        & \multirow{2}{*}{Claude Haiku 4.5}
        & GraphMemory
        & 0.75
        & 0.80 (700)
        & 0.75
        & \textbf{$\sim$57M}
        & \textbf{$\sim$9M}
        & \textbf{\$68.40} \\

        &
        & ACE (80k cap)
        & 0.75
        & \textbf{0.80 (200)}
        & \textbf{0.78}
        & $\sim$327M
        & $\sim$42M
        & \$392.40 \\

        \cmidrule(l){2-9}

        & \multirow{2}{*}{Qwen3.8-27B}
        & GraphMemory
        & 0.72
        & 0.79 (900)
        & 0.77
        & $\sim$ 31M 
        & $\sim$ 5M
        & $\$19.38$\\

        &
      & \\

        \bottomrule
    \end{tabular}%
    
    }
    \par\vspace{3pt}
\begin{minipage}{\textwidth}
    \footnotesize\raggedright
    \textsuperscript{*}Standard input/output prices per million tokens:
    Claude Haiku 4.5, \$1.00/\$5.00
    (\href{https://platform.claude.com/docs/en/about-claude/pricing}{Anthropic});
    Qwen3.8-27B, \$0.50/\$3.00
    (\href{https://www.alibabacloud.com/help/en/model-studio/model-pricing}{Alibaba Cloud Model Studio}, International tier).
   Qwen costs are API-equivalent estimates; the experiments ran locally using a quantized model.

\textsuperscript{$\dagger$}ACE/Qwen3.8-27B on \textsc{FiNER} exceeded the 48\,GB memory available on our local hardware (RTX 6000) as its full-memory context grew, even with 4-bit quantization. This highlights a practical limitation of ACE's context growth. This is preliminary workshop work, we leave a hardware-matched run to future evaluation.

\end{minipage}
\end{table*}

\begin{figure*}[t]
  \centering
  \begin{minipage}[t]{0.48\linewidth}
    \centering
    \includegraphics[width=\linewidth]{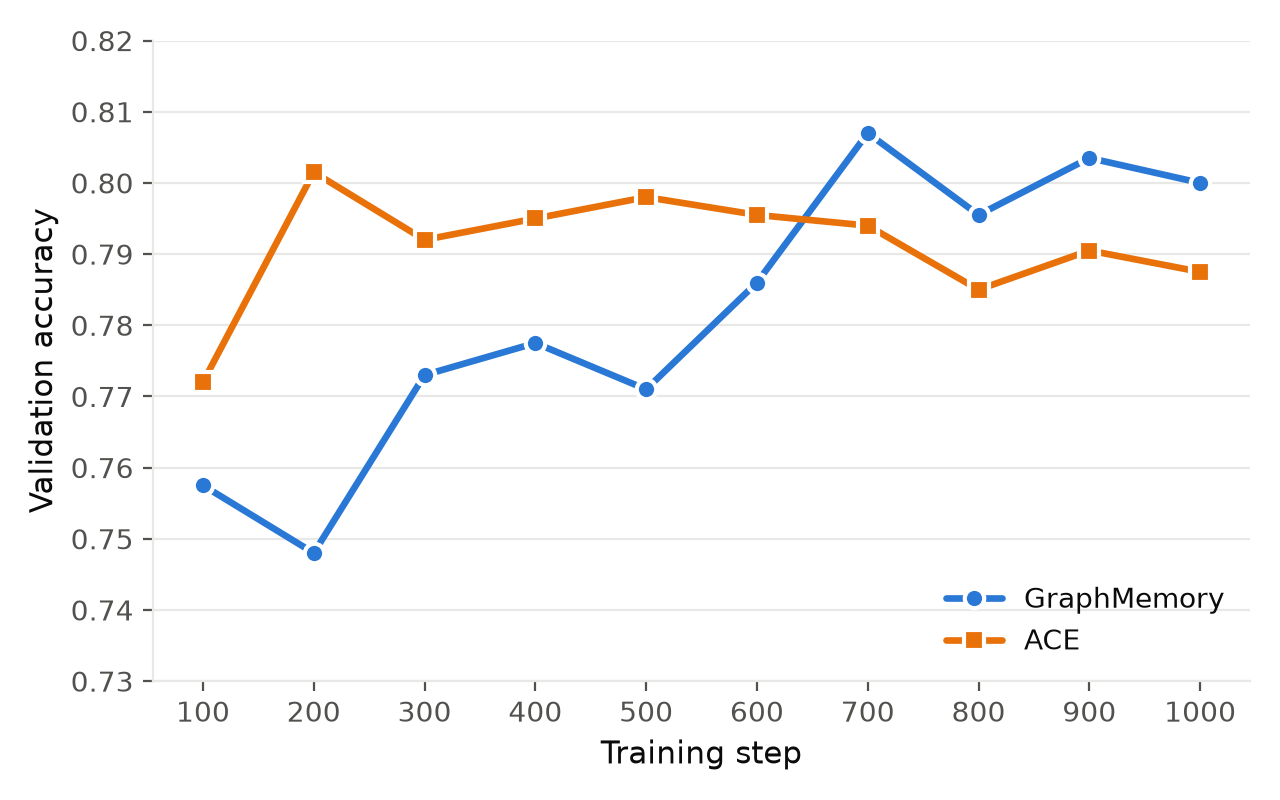}
  \end{minipage}
  \hfill
  \begin{minipage}[t]{0.48\linewidth}
    \centering
    \includegraphics[width=\linewidth]{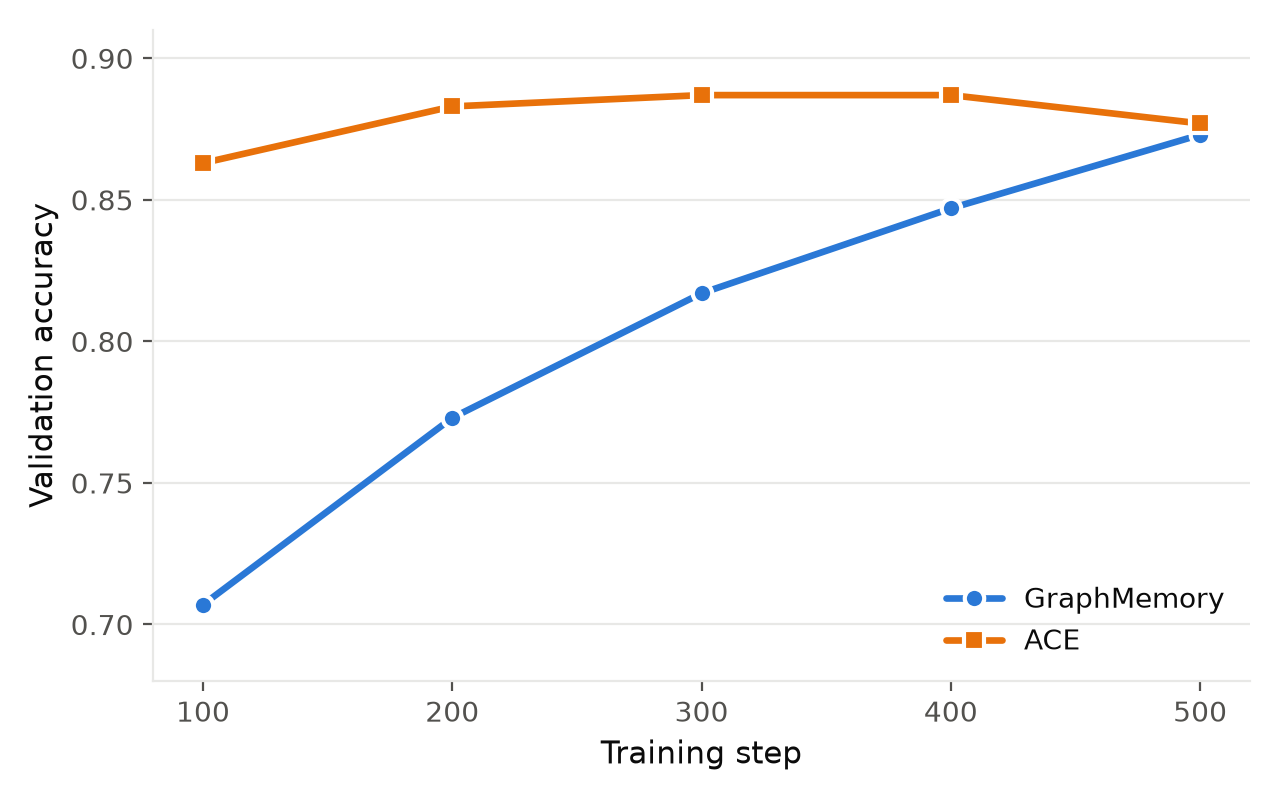}
  \end{minipage}
  \caption{
    Validation performance of GraphMemory and ACE on Claude across training steps on
    \textsc{FiNER} (left) and \textsc{Formula} (right).
  }
  \label{fig:validation_curves}
\end{figure*}

\textbf{Baselines} We compare GraphMemory against ACE \cite{zhang2026agenticcontextengineeringevolving} using two state-of-the-art LLM backbones across two financial reasoning tasks. For each backbone and task, we report cold-start performance and final test performance after memory construction. To assess adaptation efficiency, we track cumulative token consumption over the full training run and estimate the monetary cost of constructing each memory.

\textbf{Models} We evaluate GraphMemory and ACE using Claude Haiku 4.5 \cite{anthropic2025claudehaiku45} through LiteLLM and a 4-bit quantized Qwen3.8-27B \cite{qwen38} on an RTX 6000 GPU with 48\,GB of memory. For each configuration, we select the memory state that achieves the highest validation accuracy and report its performance on the held-out test set. In the Claude Haiku 4.5 experiments, we cap the ACE memory at 80,000 tokens because its growth would otherwise exceed the model's context-window limit. No such cap is required for Qwen3.8-27B, which supports a larger context window. This highlights the importance of retrieval-based memory: using ACE with smaller context windows may require a memory cap or other context-management techniques.   

\textbf{Datasets} Both memory systems are evaluated on the two financial reasoning tasks: \textsc{FiNER} \cite{loukas2022FiNER} and \textsc{Formula} \cite{wang2025finlora}. Each system is trained for a single epoch. Our \textsc{Formula} split uses 500/300/200 training, validation, and test examples, respectively, while our \textsc{FiNER} split uses 1,000/500/441 examples. We evaluate validation performance every 100 training iterations.

\textbf{Adaptation performance}
Tables~\ref{tab:combined-results} and~\ref{tab:FiNER-time} summarize our findings. Across three dataset--model configurations, GraphMemory reduces memory-construction tokens by approximately $81$--$85\%$, estimated cost by a similar margin, and test tokens by $78.6\%$--$92.2\%$, while maintaining competitive accuracy. On \textsc{Formula} with Qwen3.8-27B, it slightly outperforms ACE using only $19\%$ of its training-token budget. With Claude Haiku 4.5, ACE leads by 4.0 percentage points but uses $6.6\times$ as many training tokens. On \textsc{FiNER}, GraphMemory trails ACE by 2.94 points. Figure~\ref{fig:validation_curves} shows that ACE improves rapidly before plateauing or declining as its memory approaches the 80k-token cap. GraphMemory improves more slowly and varies more initially, but continues improving as its graph grows, suggesting a trade-off between early sample efficiency and late-stage stability.

\textbf{Token and monetary efficiency}
The largest reduction occurs on \textsc{Formula} with Claude Haiku 4.5, where GraphMemory uses approximately 20M training tokens compared with 131M for ACE, an $84.7\%$ reduction. Its test-time token usage decreases from approximately 18M to 1.4M, a $92.2\%$ reduction. Comparable savings are observed with Qwen3.8-27B, for which GraphMemory reduces training-token usage from 42M to 8M, and on \textsc{FiNER}, where usage falls from 327M to 57M. Under our pricing assumptions, these reductions lower the estimated cost of memory construction from \$157.20 to \$24.00 on Claude/\textsc{Formula}, from \$26.25 to \$5.00 on Qwen/\textsc{Formula}, and from \$392.40 to \$68.40 on Claude/\textsc{FiNER}.

\textbf{Wall-clock efficiency}
As shown in Table~\ref{tab:FiNER-time}, GraphMemory also reduces active training time. Training time decreases by $21.0\%$ on \textsc{Formula} (6.4 versus 8.1 hours) and $34.3\%$ on \textsc{FiNER} (19.0 versus 28.9 hours). GraphMemory is slower during inference, however, requiring approximately 1.8$\times$ as much time per test example. This overhead arises because GraphMemory performs retrieval before generating each prediction, whereas ACE directly supplies its complete memory to the backbone. A more efficient implementation of $\phi$, such as an embedding-based retriever, could reduce inference latency and further decrease training time. We leave this investigation to future work. Thus, GraphMemory trades a modest increase in per-query latency for substantially lower context length, training cost, and test-time token consumption.

\section{Conclusion, Limitations, and Future Work}
\vspace{-0.9em}

We cast agentic memory construction as iterative optimization over textual context, enabling a common analysis of flat and retrieval-based memory. We introduce \textsc{GraphMemory}, which organizes ACE-style procedural knowledge as a graph and retrieves bounded, query-relevant context. Related memory systems include A-MEM, which links episodic notes in a Zettelkasten-inspired structure \cite{xu2026mem}; RIZZ, which organizes procedural rules into dynamic branches \cite{goel2026rizz}; and HAGE, which learns graph traversal policies \cite{jiang2026hage}. GraphMemory learns relations between strategies from confidence-weighted helpful and harmful feedback, supporting their composition without requiring predefined edge types or training a retrieval policy. Our formulation shows how bounded retrieval changes memory-context scaling from linear growth under append-only prompting to a constant bound on retrieved content per query. Across two financial tasks and two LLM backbones, GraphMemory remains competitive with ACE while reducing memory-construction tokens by $81$--$85\%$, test-time tokens by $79$--$92\%$, and estimated API cost similarly. Learning trajectories suggest slower initial adaptation but greater stability and continued improvement.

Our results remain preliminary, covering two related tasks, two models, and one primary baseline. Broader evaluations, additional memory and prompt-optimization baselines, and repeated runs are needed. Retrieval overhead also makes individual test queries slower despite lower token usage, motivating embedding-based and hybrid retrievers, caching, and improved graph routing. Finally, longer continual-learning streams and cross-model transfer will test whether improvements persist beyond one epoch and whether memory built by larger models can efficiently adapt smaller ones.

\begin{table*}[t]
\centering
\caption{
    Active wall-clock time for offline training and held-out test on
    \textsc{Formula} and \textsc{FiNER}.
    All figures are active wall clock. Bold indicates the faster result.
}
\label{tab:FiNER-time}
\small
\setlength{\tabcolsep}{6pt}
\begin{tabular}{@{}lll cc cc@{}}
\toprule
\textbf{Dataset}
& \textbf{Model}
& \textbf{System}
& \multicolumn{2}{c}{\textbf{Training}}
& \multicolumn{2}{c}{\textbf{Final test}} \\
\cmidrule(lr){4-5}
\cmidrule(l){6-7}
&
&
& \textbf{Total}
& \textbf{Per step}
& \textbf{Total}
& \textbf{Per sample} \\
\midrule

\multirow{2}{*}{\textsc{Formula}}
& \multirow{2}{*}{claude-haiku-4-5}
& GraphMemory
& \textbf{6.4\,h}  & \textbf{46\,s}
& 15.7\,min        & 4.7\,s \\

&
& ACE (80k cap)
& 8.1\,h           & 58\,s
& \textbf{8.6\,min} & \textbf{2.6\,s} \\
\midrule

\multirow{2}{*}{\textsc{FiNER}}
& \multirow{2}{*}{claude-haiku-4-5}
& GraphMemory
& \textbf{19.0\,h} & \textbf{68\,s}
& 47.1\,min        & 6.4\,s \\

&
& ACE (80k cap)
& 28.9\,h          & 104\,s
& \textbf{26.5\,min} & \textbf{3.6\,s} \\

\bottomrule
\end{tabular}
\end{table*}

\newpage
\bibliographystyle{plainnat}
\bibliography{references}

\newpage
\appendix


\section{Additional Formalization}
\label{app:additional-formalization}

\subsection{Textual Optimization Oracles}
\label{app:optimization-oracles}

Equation~\eqref{eq:finite-context-search} suppresses how the candidate pool is
constructed. MIPROv2 searches over candidate instructions and demonstrations
for each program module. If $\mathcal I_j$ and $\mathcal E_j$ denote the
instruction and demonstration candidates for module $j$, respectively, its
search space can be abstracted as
\begin{equation}
    \mathcal Z
    = \prod_{j=1}^{m}
    \bigl(\mathcal I_j\times\mathcal E_j\bigr),
    \qquad
    \widehat z_T
    \in \arg\min_{z\in\mathcal Z_T}
    L_{\mathcal D}\bigl(c(z)\bigr),
    \label{eq:miprov2-search}
\end{equation}
where $\mathcal Z_T\subseteq\mathcal Z$ contains the configurations evaluated
within budget $T$, and $c(z)$ denotes their induced context
\cite{opsahlong2024optimizinginstructionsdemonstrationsmultistage}.

GEPA constructs its candidate set adaptively rather than drawing it from a
fully specified product space. Starting from an archive $\mathcal A_0$, it
uses execution traces and feedback to reflect on selected candidates and add
new proposals:
\begin{equation}
    \mathcal A_{t+1}
    = \mathcal A_t\cup
    \operatorname{Propose}(c_t,\tau_t),
    \qquad
    \widehat c_T
    \in \arg\min_{c\in\mathcal A_T}
     L_{\mathcal D}(c).
    \label{eq:gepa-search}
\end{equation}
Here $\operatorname{Propose}$ abstracts GEPA's reflective mutation and
candidate combination operations
\cite{agrawal2026gepareflectivepromptevolution}. Thus, $\mathcal P_T$ in
Equation~\eqref{eq:finite-context-search} may be either a searched subset of a
larger predefined space, as in MIPROv2, or an archive grown during
optimization, as in GEPA. In both cases, we abstract the resulting optimized
program or prompt as a context $c$ that is fixed during test-time inference.

A second family uses Equation~\eqref{eq:textual-context-update} to propose
textual changes iteratively. OPRO conditions the optimizer on previously
proposed contexts and their scores. TextGrad generates textual feedback and
propagates it through the computation that produced the output. TRACE uses
output feedback together with execution traces to localize the part of the
context to be modified
\cite{yang2024large, yuksekgonul2024textgrad,cheng2024trace}. When restricted
to context optimization, these methods can be viewed as different
realizations of $\mathcal O$. The gradient-descent analogy concerns the shared
update structure, not differentiability or a convergence guarantee. In
particular, $\Delta c_t$ is only a proposed direction unless an evaluation or
acceptance step verifies that it reduces the objective.

\subsection{Why Use a Local Context-Block Surrogate?}
\label{app:local-context-objective}

Allowing a separate $\Delta_i$ for each example enlarges the search space and
can produce a maximally biased solution: an unconstrained block could simply
encode the answer to its source input. Equation~\eqref{eq:context-block-risk}
is therefore not a reformulation with the same minimizer as
Equation~\eqref{eq:context-risk}. It is a surrogate for constructing the
elements of a reusable memory.

The motivation is the asymmetry between classical optimization and textual
optimization. If a reliable descent direction for the global context were
available, decomposing the problem into per-example variables would offer
little benefit and would create an obvious generalization problem. In the
black-box setting, however, the update direction must itself be produced by an
imperfect model-based oracle. Asking that oracle to extract one short lesson
from one observed success or failure is a more localized generation task than
asking it to rewrite a growing global context to improve average performance
over all examples.

Generalization is delegated to the memory interface. Given a memory bank
$\delta$, a composer or retriever constructs
$c_{\phi}(x)=\phi(\delta,x)$ for a new input. A block is useful only when it
encodes reusable information and $\phi$ exposes it on inputs for which that
information is relevant. This is why context-block quality cannot be judged
only by Equation~\eqref{eq:context-block-risk}; final validation and test risk
must still be measured using the composed or retrieved contexts, as in
Equation~\eqref{eq:graph-risk} for GraphMemory.

\subsection{Memory-Context Scaling}
\label{app:context-scaling-proofs}

Let $L_0=\operatorname{len}(c_{\delta}^{(0)})$ and
$b_i=\operatorname{len}(\Delta_i)$. Under the append-only update in
Equation~\eqref{eq:append-only-memory},
\begin{equation}
\begin{aligned}
    \operatorname{len}\left(c_{\delta}^{(i)}\right)
    &= \operatorname{len}\left(c_{\delta}^{(i-1)}\right)+b_i,\\
    \operatorname{len}\left(c_{\delta}^{(N)}\right)
    &= L_0+\sum_{i=1}^{N}b_i
    \leq L_0+NB
    = O(N).
\end{aligned}
\label{eq:append-only-recurrence}
\end{equation}
If the blocks have positive average length bounded away from zero, the final
memory-context length is $\Theta(N)$.

The cumulative number of memory-context tokens exposed across the $N$
updates obeys
\begin{equation}
\begin{aligned}
    \sum_{i=1}^{N}
    \operatorname{len}\left(c_{\delta}^{(i)}\right)
    &= NL_0+\sum_{j=1}^{N}(N-j+1)b_j\\
    &\leq NL_0+\frac{B N(N+1)}{2}
    = O(N^2).
\end{aligned}
\label{eq:append-only-cumulative-tokens}
\end{equation}
This is a bound on memory tokens supplied to the model, not on all prompt and
completion tokens used by the complete system.

For retrieval-based memory, each context is constructed afresh and does not
contain the context retrieved for the preceding example. If
$\operatorname{len}(\Delta_i)\leq B$ and at most $K$ blocks are retrieved,
then for every $i$,
\begin{equation}
    \operatorname{len}\left(c_{\phi}^{(i)}(x_i)\right)
    \leq KB = O(1),
    \qquad
    \sum_{i=1}^{N}
    \operatorname{len}\left(c_{\phi}^{(i)}(x_i)\right)
    \leq NKB
    = O(N).
    \label{eq:retrieval-cumulative-tokens}
\end{equation}
Consequently, the per-query memory context is $O(1)$. Across $N$ examples,
the cumulative memory-context exposure is $O(N)$. These bounds exclude the
stored graph, condensed index, and any context consumed internally by $\phi$.

\subsection{GraphMemory as Structured Retrieval}
\label{app:graph-formalization}

At step $t$, GraphMemory stores the blocks in
$G^{(t)}=(V^{(t)},E^{(t)})$. A new node $v^{(t)}$ stores
$\Delta^{(t)}$, while $R^{(t)}$ contains the relations introduced or updated
for that node. Under the same append-only simplification as
Equation~\eqref{eq:append-only-memory},
\begin{equation}
    V^{(t+1)}=V^{(t)}\cup\{v^{(t)}\},
    \qquad
    E^{(t+1)}=E^{(t)}\cup R^{(t)}.
    \label{eq:graph-update}
\end{equation}
The implementation may additionally edit or merge nodes and revise edge
scores; those operations do not alter the retrieval interface below.

For an input $x$, the retriever selects a query-dependent subset of nodes and
serializes their blocks:
\begin{equation}
\begin{aligned}
    S^{(t)}(x)
    &= \phi\bigl(G^{(t)},x\bigr)\subseteq V^{(t)},\\
    c_G^{(t)}(x)
    &= \bigoplus_{v_j\in S^{(t)}(x)}\Delta_j.
\end{aligned}
\label{eq:graph-retrieval}
\end{equation}
For fixed $\phi$, the empirical risk of the graph-induced contexts is
\begin{equation}
   L_{\mathcal D}^{\mathrm{graph}}\bigl(G^{(t)}\bigr)
    = \frac{1}{n}\sum_{i=1}^{n}
    \mathbb E\left[
        \ell\bigl(f(c_G^{(t)}(x_i),x_i),y_i\bigr)
    \right].
    \label{eq:graph-risk}
\end{equation}
GraphMemory constructs $G^{(0)},G^{(1)},\ldots$ with the aim of reducing this
risk. At the level of memory exposure, supplying the full memory is recovered
as the special case $S^{(t)}(x)=V^{(t)}$ for every $x$. GraphMemory instead
uses a bounded, query-dependent set, so a specialized block affects only the
inputs for which it is retrieved.

\section{Implementation Details}
\label{app:implementation-details}

\subsection{Directed Evidence-Weighted Memory Graph}

GraphMemory maintains a directed graph
$G^{(t)}=(V^{(t)},E^{(t)})$. Each node stores a unique identifier, a short
routing key, and the full ACE-style bullet point used as context. The routing
index contains only the identifier and short key; full bullet-point content is
loaded only after graph retrieval. A directed edge $e=(u,v)$ records that the
knowledge in node $v$ may be useful after node $u$ has been retrieved. The
implementation does not store a single mutable weight. Instead, each directed
edge stores two nonnegative, confidence-weighted evidence counters:
$S_e^{(t)}$ for helpful transitions and $F_e^{(t)}$ for harmful transitions.
If edge $e$ is created at step $\tau_e$, its counters are initialized as
\begin{equation}
    S_e^{(\tau_e)}=F_e^{(\tau_e)}=0.
    \label{eq:edge-counter-initialization}
\end{equation}
We treat these counters as fractional pseudo-counts under a uniform
$\operatorname{Beta}(1,1)$ prior. Under this fractional-observation
interpretation, the edge parameter is
\begin{equation}
    \theta_e \mid \mathcal H_t
    \sim \operatorname{Beta}\!\left(1+S_e^{(t)},\,1+F_e^{(t)}\right),
    \label{eq:edge-beta-posterior}
\end{equation}
where $\mathcal H_t$ is the evidence accumulated through step $t$. The value
used to rank the edge during retrieval is the posterior mean
\begin{equation}
    w_e^{(t)}
    = \mathbb E[\theta_e\mid\mathcal H_t]
    = \frac{S_e^{(t)}+1}{S_e^{(t)}+F_e^{(t)}+2}.
    \label{eq:edge-retrieval-weight}
\end{equation}
Thus, a new edge has neutral weight $w_e^{(\tau_e)}=1/2$ even though no
explicit weight field is stored. Because the graph is directed, $(u,v)$ and
$(v,u)$ have separate counters and can receive different retrieval weights.

After generation, the Generator reports an ordered influence trace
$M_t=[n_1,\ldots,n_k]$ containing the loaded memory nodes that it attributes as
influential to its answer. Only identifiers among the nodes loaded for that
example are accepted. Adjacent items define candidate directed transitions,
\begin{equation}
    \mathcal T(M_t)
    = \{(n_1,n_2),(n_2,n_3),\ldots,(n_{k-1},n_k)\}.
    \label{eq:generator-transition-trace}
\end{equation}
The Reflector labels each candidate as helpful, harmful, or neutral and returns
an evidence strength $c_e^{(t)}\in[0,1]$. For an existing edge $e$, the evidence
update is
\begin{equation}
\left(S_e^{(t+1)},F_e^{(t+1)}\right)=
\begin{cases}
\left(S_e^{(t)}+c_e^{(t)},F_e^{(t)}\right),
    & \text{if $e$ is helpful},\\
\left(S_e^{(t)},F_e^{(t)}+c_e^{(t)}\right),
    & \text{if $e$ is harmful},\\
\left(S_e^{(t)},F_e^{(t)}\right),
    & \text{if $e$ is neutral}.
\end{cases}
\label{eq:edge-evidence-update}
\end{equation}
The confidence value therefore controls the fractional amount of evidence
contributed by one observation; neutral feedback and $c_e^{(t)}=0$ leave both
counters unchanged, as do edges that were not observed on the example.
Hallucinated or unloaded node identifiers, duplicate
pairs, invalid labels, and confidences outside $[0,1]$ are discarded before
this update, so each unique transition is updated at most once per example.
Observing a transition does not implicitly create an edge: new edges can be
introduced only by validated Curator operations, with both counters initialized
to zero. The Curator may likewise issue a separate validated edge-removal
operation when a relation should no longer be retained; deletion is not an
automatic consequence of Equation~\eqref{eq:edge-retrieval-weight}. Node
additions and revisions are also local, and new nodes are inserted into the
routing index without rewriting the remainder of the graph.

The retrieval weight is a smoothed estimate of how often the Reflector has
judged an observed directed transition helpful under the current retrieval
policy. It should not be interpreted as a causal measure of interaction between
two nodes or as a calibrated probability that retrieving the edge will improve
accuracy. Evidence is policy-dependent because highly ranked edges are more
likely to be observed again, and ranking by the posterior mean does not
explicitly reward uncertainty. These limitations are important when
interpreting the learned graph.


\subsection{Retrieval and Traversal}

Retrieval has two stages. First, an LLM Retriever reads the compact index and
returns a structured selection of up to two index sections and at most six
entry nodes. Second, GraphMemory performs a weight-prioritized breadth-first
traversal over outgoing edges. Entry nodes have depth zero. At each breadth
level, active outgoing neighbors are ordered by decreasing posterior-mean
weight from Equation~\eqref{eq:edge-retrieval-weight}, with target-node
identifier used as a deterministic tie-break, and at most the two
highest-scoring neighbors of a loaded node are considered. Traversal
is bounded by two hops from each entry node, at most 12 loaded nodes, and at
most 2,500 tokens of serialized node content. A node is not loaded if doing so
would violate either the node or token budget.

Algorithm~\ref{alg:GraphMemory-retrieval} gives the retrieval procedure. The
queue preserves breadth-first depth order, while edge scores determine the
priority among neighbors at the same expansion point. Duplicate nodes reached
from different entries are loaded only once.

\begin{algorithm}[t]
\caption{Bounded posterior-mean-prioritized retrieval in GraphMemory}
\label{alg:GraphMemory-retrieval}
\begin{algorithmic}[1]
\REQUIRE Query $x$, directed graph $G=(V,E)$, compact index $I(G)$
\STATE $A \leftarrow \operatorname{LLMSelect}(I(G),x)$
       \COMMENT{at most 2 sections and 6 valid entry nodes}
\STATE $Q \leftarrow [(v,0):v\in A]$, $L\leftarrow[\,]$, $Z\leftarrow\varnothing$
\WHILE{$Q\neq\varnothing$ and $|L|<12$}
    \STATE $(v,d)\leftarrow\operatorname{popFront}(Q)$
    \IF{$v\notin Z$}
        \STATE $Z\leftarrow Z\cup\{v\}$
        \IF{$\operatorname{len}(\operatorname{Ser}(L\mathbin{\|}[v]))\leq 2500$}
            \STATE append $v$ to $L$
            \IF{$d<2$}
                \STATE $N_v\leftarrow\operatorname{Top2}_{u:(v,u)\in E}w_{(v,u)}$
                \STATE append $[(u,d+1):u\in N_v]$ to $Q$
            \ENDIF
        \ENDIF
    \ENDIF
\ENDWHILE
\STATE \textbf{return} $\operatorname{Ser}(L)$
\end{algorithmic}
\end{algorithm}

The Retriever uses structured JSON output. Invalid identifiers are discarded.
We use no full-memory fallback: if the graph is empty or no valid entry node is
selected, the retrieved memory is empty and generation proceeds without a
memory block. This choice prevents retrieval failures from silently reverting
to the full-playbook exposure that GraphMemory is designed to avoid.

For each retrieval, the implementation records the selected section and entry
identifiers, loaded node identifiers, considered and traversed edges, loaded
memory tokens, maximum hop depth, traversal time, and whether a fallback was
used. These traces both support token and latency accounting and restrict
relation updates to content that was actually available during generation.

\subsection{Retrieve--Generate--Reflect--Curate Execution Order}

For training example $(x_t,y_t)$, GraphMemory first retrieves from the graph
constructed from earlier examples,
\begin{equation}
    c_t=\phi(G^{(t)},x_t),
    \qquad
    \widehat y_t=f(c_t,x_t).
    \label{eq:implementation-predict-order}
\end{equation}
Only after producing $\widehat y_t$ does the Reflector inspect the prediction,
available feedback, and execution trace. The Curator then emits local graph
edits, which are applied to obtain $G^{(t+1)}$. Consequently, information
derived from $y_t$ cannot affect the prediction on $x_t$; it becomes available
only to subsequent examples. The Generator and Reflector retain ACE's core roles, with the additional
memory-use tracing and edge feedback described above, while the Retriever and graph-valued Curator
implement the two GraphMemory-specific interfaces described in
Section~\ref{sec:GraphMemory}.

\subsection{Model and Evaluation Configuration}

For the Qwen3.8-27B experiments, the model is loaded through Hugging Face. We
retain the checkpoint configuration except for three explicit runtime choices:
4-bit quantization, scaled dot-product attention (SDPA), and disabled thinking
mode. Disabling thinking prevents unnecessarily long reasoning outputs. These
model and inference settings are identical for ACE and GraphMemory. For the
Claude Haiku 4.5 experiments, all model calls are made through LiteLLM using
its default request configuration. Within each dataset--backbone comparison,
ACE and GraphMemory use the same backbone and model configuration; the
method-specific difference is the memory interface rather than the underlying
language model.


Each system processes the training split for one epoch, and the then-current
memory is evaluated on the validation split every 100 updates. The checkpoint
with the highest validation accuracy is selected for the held-out test
evaluation. In the Claude experiments, ACE uses the 80,000-token memory cap
reported in the main text. Formula uses 500/300/200 train/validation/test
examples, and FiNER uses 1,000/500/441 examples.


\subsection{Token and Latency Accounting}

The bounded-retrieval result in Proposition~\ref{prop:bounded-retrieval}
applies to the serialized node content supplied to the Generator. It does not
include the compact index read by the Retriever. Because the present Retriever
reads the complete index, its input grows with the number of indexed nodes.
Total system usage includes the Retriever's index, retrieved node content,
fixed instructions and query tokens, and the remaining input and output tokens
from the Retriever, Generator, Reflector, and Curator. Only the retrieved
node content supplied to the Generator is covered by the constant retrieved-content bound. The wall-clock measurements
include the additional retrieval call made by GraphMemory, which explains why
GraphMemory can use fewer tokens while having higher per-query latency.

 
\newpage

\end{document}